\pdfoutput=1

\documentclass[11pt]{article}

\usepackage{acl}
\usepackage{latexsym}

\usepackage[T1]{fontenc}
\usepackage{ragged2e}
\usepackage{siunitx}
\usepackage{adjustbox}
\usepackage{amsmath,amsfonts,bm}
\usepackage{mathtools}
\newcommand*\rot{\rotatebox{90}}
\newcommand{\STAB}[1]{\begin{tabular}{@{}c@{}}\rot{#1}\end{tabular}}

\usepackage{titlesec}
\titlespacing*{\section}{0.5ex}{1ex}{1ex}
\titlespacing*{\subsection}{1ex}{1ex}{.5ex}
\titlespacing*{\subsubsection}{0pt}{.2ex}{.2ex}

\usepackage{amsfonts}

\usepackage{microtype}

\usepackage[subtle]{savetrees}
\usepackage{multirow}
\usepackage{hyperref}
\usepackage{booktabs} 
\usepackage{tabularx}
\usepackage{graphicx}
\usepackage{subcaption}
\usepackage[htt]{hyphenat}
\usepackage{sidecap}
\usepackage{times}

\usepackage[T1]{fontenc}

\usepackage[utf8]{inputenc}

\usepackage{microtype}

\title{Quantifying  Privacy Risks of Masked Language Models\\  Using  Membership Inference Attacks}

\author{Fatemehsadat Mireshghallah\textsuperscript{\rm 1}, Kartik Goyal\textsuperscript{\rm 2}, Archit Uniyal\textsuperscript{\rm 3}\\
    \textbf{Taylor Berg-Kirkpatrick}\textsuperscript{\rm 1},
   \textbf{Reza Shokri}\textsuperscript{\rm 4} \\
    \textsuperscript{\rm 1} University of California San Diego,
    \textsuperscript{\rm 2} Toyota Technological Institute at Chicago (TTIC)\\
    \textsuperscript{\rm 3} University of Virginia,
    \textsuperscript{\rm 4} National University of Singapore \\
    \texttt{[fatemeh, tberg]@ucsd.edu},\\ \texttt{ kartikgo@ttic.edu,a.uniyal@virginia.edu,reza@comp.nus.edu.sg}
  }

\begin{document}
\maketitle
\begin{abstract}
The wide adoption and application of Masked language models~(MLMs) on sensitive data (from legal to medical) necessitates a thorough quantitative investigation into their privacy vulnerabilities. 
Prior attempts at measuring leakage of MLMs via membership inference attacks have been inconclusive, implying potential robustness of MLMs to privacy attacks.
In this work, we posit that prior attempts were inconclusive because they based their attack solely on the MLM's model score. We devise a stronger membership inference attack based on likelihood ratio hypothesis testing that involves an additional reference MLM 
to more accurately quantify the privacy risks of memorization in MLMs. We show that masked language models are indeed susceptible to likelihood ratio membership inference attacks: Our empirical results, on models trained on medical notes, show that our attack improves the AUC of prior membership inference attacks from $0.66$ to an alarmingly high $0.90$ level.

\end{abstract}
\section{Introduction}

BERT-based encoders with Masked Language Modeling  (MLM) Objectives~\citep{devlin2018bert, liu2019roberta}  have become models of choice for use as pre-trained models for various Natural Language Processing (NLP) classification tasks~\citep{wang2018glue, zhang2019bertscore, rogers-etal-2020-primer} and have been applied to diverse domains such as disease diagnosis, insurance analysis on financial data, sentiment analysis for improved user experience, etc~\cite{yang2020finbert, gu2021domain, lee2020biobert}. 
Given the sensitivity of the data used to train these models, it is crucial to conceive a framework to systematically evaluate the leakage of training data from these models~\cite{shokri_talk, carlini2019secret, murakonda2020ml,mireshghallah2020privacy}, and limit the leakage. 
The conventional way to measure the leakage of training data from machine learning models is by performing membership inference attacks~\cite{shokri2017membership, nasr2021adversary}, in which the attacker tries to determine whether a given sample was part of the training data of the target model or not. These attacks expose the extent of memorization by the model at the level of individual samples. 
Prior attempts at performing membership inference and reconstruction attacks on masked language models have either been inconclusive~\cite{lehman-etal-2021-bert}, or have (wrongly) concluded that memorization of sensitive data in MLMs is very limited and these models are more private than their generative counterparts (e.g., autoregressive language models)~\cite{Vakili1633747, jagannatha2021membership, nakamura2020kart}.

We hypothesize that prior MLM attacks have been inconclusive because they rely solely 
on the target model's (model under attack) loss on each individual sample as a proxy for how well the model has memorized that sample. If the loss is lower than a threshold, the sample is predicted to be a member of the training set.   
However, the target model's loss includes confounding factors of variation like the intrinsic complexity of the sample -- and thus provides a limited discriminative signal for membership prediction. This scheme has either a high false-negative rate (with a conservative threshold) -- classifying many hard-to-fit samples from the training set as non-members, or a high false-positive rate (with a generous threshold) -- failing to identify easy-to-fit samples that are not in the training set.

Reference-based likelihood ratio attacks, on the other hand, when applied to certain probabilistic graphical models and classifiers, have been shown to alleviate this problem and more accurately distinguish members from non-members~\cite{murakonda2021quantifying,ye2021enhanced}.
In such attacks, instead of the loss of the model under attack, we look at the ratio of the likelihood of the sample under the target model and a reference model trained on samples from the underlying population distribution that generates the training data for the target model. This ratio recalibrates the test statistic to explain away spurious variation in model's loss for different samples due to the intrinsic complexity of the samples. Unlike most other models (e.g., generative models), however, computing the likelihood of MLMs is not straightforward. 
In this paper, we propose a principled framework for measuring information leakage of MLMs through likelihood ratio-based membership inference attacks and perform an extensive analysis of memorization in such models.
To compute the likelihood ratio of the samples under the target and the reference MLMs, we view the MLMs as energy-based probabilistic models~\cite{goyal2022exposing} over the sequences. This enables us to perform powerful inference attacks on conventionally non-probabilistic models like masked language models.  

We evaluate our proposed attack on a suite of masked clinical language models, following~\cite{lehman-etal-2021-bert}. We compare our attack with the baseline from the prior work that relies solely on the loss of the target model~\cite{yeom2018privacy,song2020information,jagannatha2021membership}. 
We empirically show that \textit{our attack improves the AUC from $0.66$ to $0.90$} on the ClinicalBERT-Base model, and achieves ~\textit{a true positive rate (recall) of $~79.2\%$ } (for a false positive rate of $10\%$), which is a substantial improvement over the baseline with $15.6\%$ recall.  This shows that, contrary to prior results, masked language models are significantly susceptible to attacks exploiting the leakage of their training data. In low error regions (at $1\%$ false positive rate) \textit{our attack is $51\times$ more powerful than the prior work}.

We also present analyses 
of the effect of the size of the model, the length of the samples, and the choice of the reference model on the success of the attack. Finally, we attempt to identify features of samples that are more exposed (attack is more successful on), and observe that samples with multiple non-alphanumeric symbols (like punctuation) are more prone to being memorized. We provide instructions on how to request access to the data and code in Appendix~\ref{sec:app:data}.
%

\section{Membership Inference Attacks}

In this section, we first formally describe the membership inference attack, how it can be conducted using likelihood ratio tests and how we apply the test for masked language models (MLMs) which do not explicitly offer an easy-to-compute probability distribution over sequences. Finally, we describe all the steps in our attack, as summarized in Figure~\ref{fig:proc}.


\subsection{Problem Formulation}

Let $M_\theta$ denote a model with parameters $\theta$ that have been trained on data set $D$, sampled from the general population distribution $p$. Our goal is to quantify the privacy risks of releasing $M_\theta$ for the members of training set $D$.

We consider an adversary who has access to the target model $M_{\theta}$. We assume this adversary can train a (reference) model $M_{\theta_R}$ with parameters $\theta_R$ on independently sampled data from the general population $p$.
In a Membership Inference Attack (MIA), the objective of the adversary is to create a decision rule that determines whether a given sample $s$ was used for training $M_\theta$.  To test the adversary, we perform the following experiment. We sample a datapoint $s$ from either the general population or the training data with a $0.5$ probability, and challenge the adversary to tell if $s$ is selected from the training set (it is a member) or not (it is a non-member)~\cite{murakonda2021quantifying}.
The precision of the membership inference attack indicates the degree of information leakage from the target model about the members
of its training set. We measure the adversary’s success using two metrics: (1) the adversary’s power (the true positive rate), and (2) the adversary's error (the false positive rate).

\subsection{Likelihood Ratio Test}\label{sec:lr}
Before discussing our proposed attack for MLMs in the next section, we summarize the likelihood ratio test here which forms the core of our approach. A likelihood ratio test distinguishes between a null hypothesis and an alternative hypothesis via a test statistic based on the ratio of likelihoods under the two hypotheses. Prior work demonstrated an MIA attack based on the likelihood ratio to be optimal for probabilistic graphical models (Bayesian networks)~\cite{murakonda2021quantifying}. Given a sample $s$ from the training data of the target model, the adversary aims at distinguishing between two hypotheses: 

\begin{enumerate}
\item Null hypothesis ($H_{\textrm{out}}$): The target sample~$s$ is
drawn from the general population~$p$, independently from the training set~$D$.
\item Alternative hypothesis ($H_{\textrm{in}}$): The target sample~$s$ is drawn from the target model's training set~$D$.
\end{enumerate}

\begin{figure*}[t]
    \centering
     \includegraphics[width=0.95\linewidth]{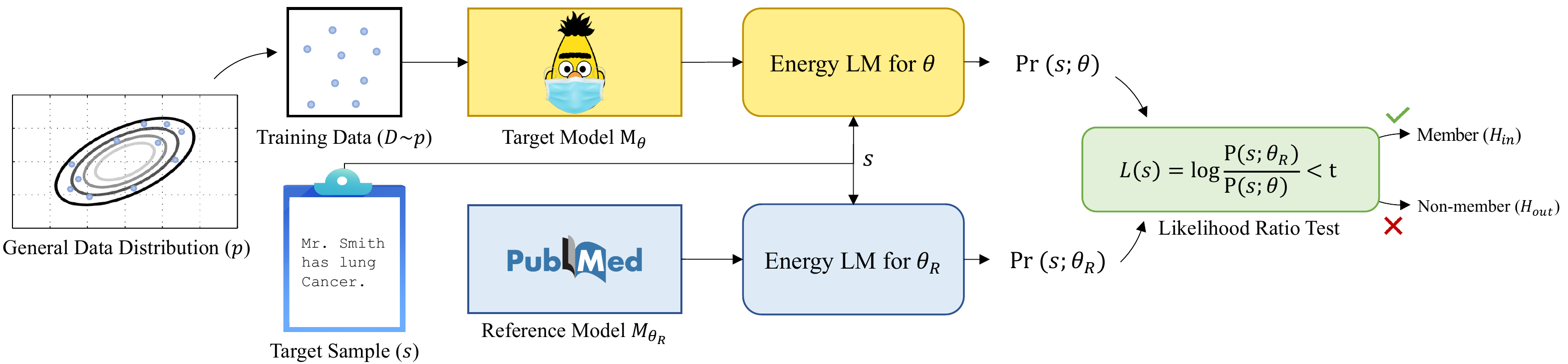}
    \caption{Overview of our  attack: to determine whether a target sample $s$ is a member of the training data ($D \sim p$) of the target model ($M_{\theta}$), we feed it to the energy function formulation of $M_{\theta}$ so that we can compute $\Pr(s;M_{\theta})$, the probability of~$s$ under $M_{\theta}$. We do the same with a reference model $M_{\theta_R}$ which is trained on a disjoint data set from the same distribution as the training data. Then, we compute likelihood ratio $L(s)$, and based on this ratio and a given test threshold $t$, we decide if $s$ is a member of $D$ ($H_{\textrm{in}}$) or not ($H_{\textrm{out}}$).}
    \label{fig:proc}
    \vspace{-2ex}
\end{figure*}

The goal of hypothesis testing is to find whether there is enough evidence to reject $H_{\textrm{out}}$ in favor of $H_{\textrm{in}}$. We use a likelihood ratio for this purpose which involves comparison of the likelihood of the target sample under the settings for $H_{\textrm{out}}$ and $H_{\textrm{in}}$ respectively. For $H_{\textrm{in}}$, we already have access to the target model, which is parameterized by $\theta$ and trained on $D$. For $H_{\textrm{out}}$, we require access to a model trained on the general population. As mentioned earlier, the adversary has access to a reference model parameterized by $\theta_R$. Therefore, the likelihood ratio test is characterized by the following statistic: 
\begin{equation}
\label{eq:lr}
 L(s) =\log\left( \frac{p(s;~\theta_R)}{p(s;~\theta)}\right)
\end{equation}
The Likelihood Ratio (LR) test is a comparison of the log-likelihood ratio statistic~$L(s)$ with a threshold~$t$. If $L(s) \leq t$, then the adversary rejects $H_{\textrm{out}}$ (decides in favor of membership of $s \in D$); otherwise the adversary fails to reject $H_{\textrm{out}}$. We discuss the details of selecting the threshold and quantifying the attack's success in Section~\ref{sec:attack}.

\subsection{Likelihood Ratio Test for MLMs}\label{sec:mlm-lr}
Performing a likelihood ratio test with masked language models is difficult because these models do not explicitly define an easy-to-compute probability distribution over natural language sequences. Following prior work~\cite{goyal2022exposing}, we alternatively view pre-trained MLMs as energy-based probability distributions on sequences, allowing us to directly apply the likelihood ratio formalism.
An energy-based sequence model defines the probability distribution over the space of possible sequences $\mathcal{S}$ as: 
\begin{align*}
    p(s;\theta) = \frac{e^{-E(s;\theta)}}{Z_\theta},
\end{align*}
where $E(s;\theta)$ refers to the scalar energy of a sequence $s$ that is parametrized by $\theta$, and $Z_\theta=\sum_{s' \in \mathcal{S}} e^{-E(s'; \theta)}$ denotes the intractable noramlization constant. Under this framework, the likelihood ratio test statistic~(Eq.~\ref{eq:lr}) is:
\begin{equation}
\begin{split}
    L(s)&=\log\left( \frac{p(s;~\theta_R)}{p(s;~\theta)}\right) \\
    &= \log\left({\frac{e^{-E(s;~\theta_R)}}{Z_{\theta_R}}}\right) - \left(\log{\frac{e^{-E(s;~\theta)}}{Z_\theta}}\right) \nonumber\\
    &= -E(s;~\theta_R)-\log(Z_{\theta_R})+E(s;~\theta)+\log(Z_\theta) \nonumber\\
    &=E(s;~\theta)-E(s;~\theta_R) \nonumber +\textrm{constant}
\end{split}
\label{eq:mlr}
\end{equation}
Above, we make use of the fact that for two fixed models (i.e., target model ~$\theta$, and reference model $\theta_R$), the intractable term $\log({Z_\theta})-\log({Z_{\theta_R}})$ is a global constant and can be ignored in the test. Therefore, computation of the test statistic only relies on the difference between the energy values assigned to sample $s$ by the target model $M_\theta$, and the reference model $M_{\theta_R}$.

In practice, we cast a traditional MLM as an energy-based language model using a slightly different parameterization than explored by \citet{goyal2022exposing}. Since the training of most MLMs (including the ones we attack in experiments) involves masking $15\%$ of the tokens in a training sequence, we define our energy parameterization on these $15\%$ chunks. Specifically, for a sequence of length $T$, and the subset size $l=\lceil 0.15\times T \rceil$, we consider computing the energy with the set $\mathcal{C}$ consisting of all ${T \choose l}$ combinations of masking patterns. 
\begin{equation}
\label{eq:energy}
    E(s;~\theta) = -\frac{1}{\lvert\mathcal{C}\rvert}\sum_{I \in \mathcal{C}}\sum_{i \in I}\log\left( p_{\textrm{mlm}}(s_i|s_{\backslash I};~\theta)\right)
\end{equation}

where $s_{\backslash I}$ is the sequence $s$ with the $l$ positions in $I$ masked. Computing this energy, which involves running $\lvert \mathcal{C}\rvert = {T \choose l}$ forward passes of the MLM, is expensive. Hence, we further approximate this parametrization by summing up over $K$ random masking patterns where $K \ll \lvert \mathcal{C} \rvert$.

\subsection{ Quantifying the Privacy Risk}\label{sec:attack}

Given the form of the likelihood ratio test statistic (Eq.~\ref{eq:mlr}) and energy function formulation for MLM likelihood (Eq.~\ref{eq:energy}), we conduct the attack as follows (shown in Figure~\ref{fig:proc}):

\begin{enumerate}
    \item Given a sample $s$ whose membership we want to determine, we calculate its energy $E(s;~\theta)$ under the model under attack ($M_\theta$) using Eq.~\ref{eq:energy}.  We calculate the energy  $E(s;~\theta_R)$ under the reference model. Using Eq.~\ref{eq:lr}, we compute the test statistic $L(s)$ by subtracting the two energies. 
    \item We compare $L(s)$ to a threshold $t$, and if $L(s) \leq t$, we reject the null hypothesis ($H_{\textrm{out}})$ and mark the sample as a member.  Otherwise, we mark it as a non-member. 
\end{enumerate}

\begin{figure}[t]
    \centering
    \begin{subfigure}{0.45\textwidth}
     \includegraphics[width=\linewidth]{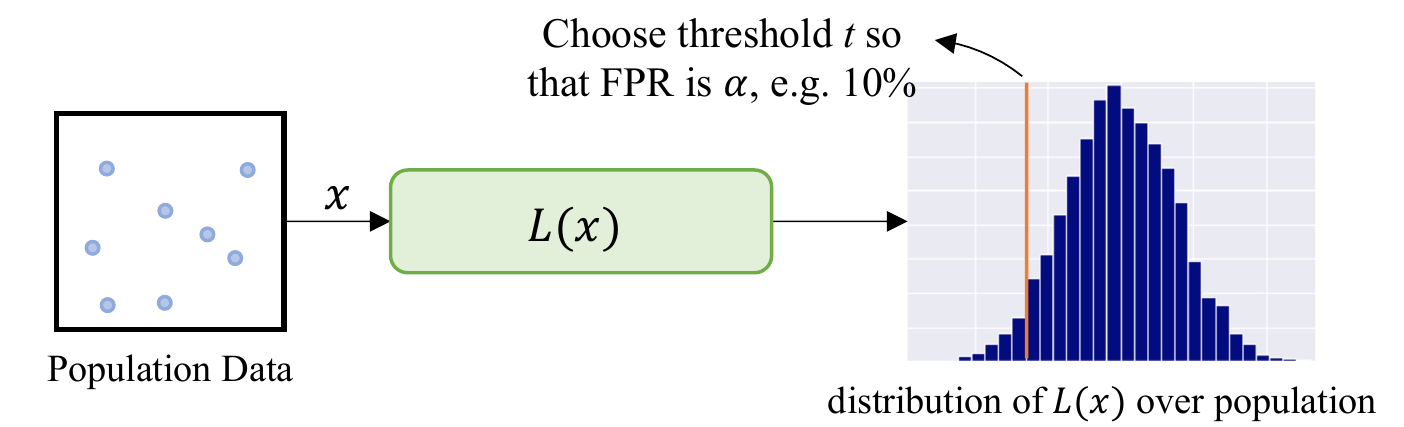}
     \footnotesize
     \caption{Selecting  threshold $t$}
     \label{fig:threshold}
    \end{subfigure}    
    ~
    \begin{subfigure}{0.45\textwidth}
     \includegraphics[width=\linewidth]{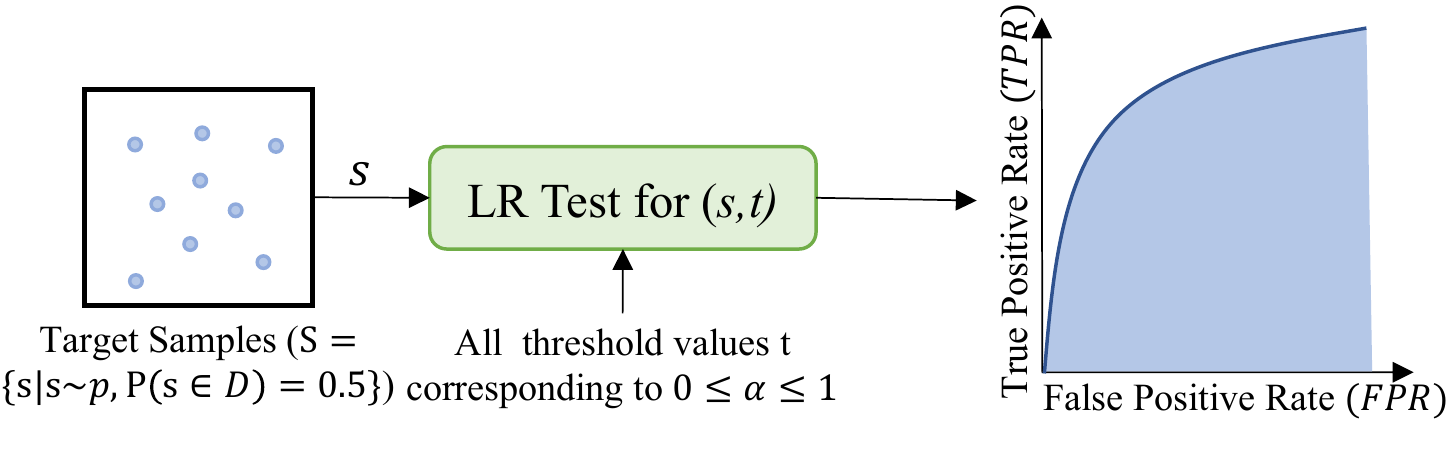}
    \footnotesize 
     \caption{Plotting the ROC curve}
     \label{fig:roc}
    \end{subfigure}
    \caption{(a) Selecting a threshold for the attack using population data and (b) plotting the ROC curve to show the true-positive vs. false-positive rate trade-off, given different thresholds. } 
    \label{fig:eval}
    \vspace{-2ex}
\end{figure}

\noindent\textbf{Choosing the threshold.} The threshold determines the (false positive) error the adversary is willing to tolerate in the membership inference attack. Thus, for determining the threshold $t$, we select a false positive rate $\alpha$, and empirically compute $t$ as the corresponding percentile of the likelihood ratio statistic over random samples from the underlying distribution. 
This process is visualized in Figure~\ref{fig:threshold}. We empirically estimate the distribution of the test statistic~$L(x)$ using all the sequences $x$ drawn from the general population distribution. This yields the distribution of~$L$ under the null hypothesis.  
We then select the threshold such that the tolerance of attack's error i.e. the rate at which attack \textit{falsely} classifies the population data as ``members'' is $\alpha\%$.

\noindent\textbf{Quantifying the Privacy Risk.} The attacker's success (i.e. the privacy loss of the model) can be quantified using the relation between the attack's power (the true positive rate) versus its error (the false positive rate).  Higher power for lower errors indicates larger privacy loss.
To compare two attack algorithms (e.g., our method versus the target model loss based methods), we can compute their power for all different error values, which can be illustrated in an ROC curve (as in Figure~\ref{fig:roc} and Figure~\ref{fig:result:roc}).  This enables a complete comparison between two attack algorithms.  The Area Under the Curve (AUC) metric for each attack provides an overall threshold independent evaluation of the privacy loss under each attack.

\section{Experimental Setup}

We conduct our experiments using the pre-processed data, and pre-trained models provided by~\citet{lehman-etal-2021-bert}.
We use this medical-based setup as medical notes are sensitive and leakage of models trained on notes can cause privacy breaches. In this section, we briefly explain the details of our experimental setup. Appendix~\ref{app:setup} provides more details. Table~\ref{tab:experiment_notation} provides a summary. 
%

\subsection{Datasets}

We run our attack on two sets of target samples, in both of which the ``members'' portion is sampled from the training set ($D$) of our target models, which is the MIMIC-III dataset. The non-members, however,  are different. For the results shown under ``MIMIC'', the non-members are a held-out subset of the MIMIC data that was not used in training. For i2b2, the non-members are from a different (but similar) dataset, i2b2. Below we elaborate on each of these datasets (full detail in Appendix~\ref{app:datasets}). Both the datasets require a license for access, so we cannot show examples of the training data.

\noindent\textbf{MIMIC-III.} The target models we attack are trained on the pseudo re-identified MIMIC-III notes which consist of $1,247,291$  electronic health records (EHR) of  $46,520$ patients.
%


\noindent\textbf{i2b2.} This dataset was curated for the i2b2 de-identification of protected health information (PHI) challenge in 2014~\cite{i2b2}. 
We use this dataset as a secondary non-member dataset since it is similar in domain to MIMIC-III (both are medical notes), is larger in terms of size than the held-out MIMIC-III set,  and has not been used as training data for our models.
%


%

\begin{table*}[]
    \centering
    \caption{Summary of model and baseline notations used in the results.}
    \vspace{-2ex}
    \label{tab:experiment_notation}
    \begin{adjustbox}{width=0.92
    \linewidth, center}

\begin{tabular}{cllll}
	\toprule
	&Notation &  {Explanation} \\
    \midrule
    \multirow{4}{*}{\STAB{Models}} 
&   Base	&   ClinicalBERT-base target model, trained for 300k iterations w/ sequence length 128 and  100k iterations w/ sequence length 512.   \\
&   Base++	&   ClinicalBERT++ target model, same as the Base model but trained for longer: trained for 1M iterations w/ a sequence length of 128. \\
&   Large	&   ClinicalBERT-large target model, trained for 300k iterations w/ sequence length 128 and  100k iterations w/ sequence length 512.   \\
&   Large++	&   ClinicalBERT-large++ target model, same as the Large model but trained for longer: trained for 1M iterations w/ a sequence length of 128. \\
      \midrule[0.1pt]
\multirow{3}{*}{\STAB{Methods}} 
&   (A) w/ $\mu$ thresh.    &	Baseline with threshold set to be the mean of training sample losses ($\mu$)  (for reporting threshold-dependant metrics)  \\
&   (A) w/ Pop. thresh. &	Baseline with threshold set so that there is $10\%$ false positive rate (for reporting threshold-dependant metrics) \\
&   (B) w/ Pop. thresh. &	Our method with threshold set so that there is $10\%$ false positive rate  ( for reporting threshold-dependant metrics)  \\
	\bottomrule
\end{tabular}

    \end{adjustbox}
    \vspace{-2ex}
\end{table*}

\subsection{ Models}

\noindent\textbf{Target Models.}
We perform our attack on $4$ different pre-trained ClinicalBERT models, that are all trained on MIMIC-III, but with different training procedures, summarized in Table~\ref{tab:experiment_notation} under Models.






\noindent\textbf{Reference Models.} We use  \texttt{Pubmed-BERT} \footnote{\url{bionlp/bluebert\_pubmed\_uncased\_L-12\_H-768\_A-12}} trained on pre-processed PubMed texts containing around 4000M words extracted from  PubMed ASCII code version~\cite{peng2019transfer} as our main domain-specific reference model, since its training data is similar to MIMIC-III in terms of domain, however, it does not include MIMIC-III training data. We also use the standard pre-trained \texttt{bert-base-uncased} as a general-domain reference model for ablating our attack. 


\subsection{Baselines}

We compare our results with a popular prior method, which uses the loss of the target model as a signal to predict membership~\cite{yeom2018privacy,jayaraman2021revisiting,ye2021enhanced}. We show this baseline as \textit{Model loss} in our tables. This baseline could have two variations, based on the way its threshold is chosen: (1) \textit{$\mu$ threshold}~\cite{jagannatha2021membership}, which assumes access to the mean of the training data loss, $\mu$ and uses it as the threshold for the attack, and (2) \textit{ population threshold (pop. thresh.)} which calculates the loss on a population set of samples (samples that were not used in training but are similar to training data), and then selects the threshold that would result in a  $10\%$ false positive rate on that population.

\subsection{Metrics}

\noindent\textbf{Area Under the ROC Curve (AUC).}
The ROC curve is a plot of power (true positive rate) versus error (false positive rate), measured across different thresholds $t$, which captures the trade-off between power and error. Thus, the area under the ROC curve (AUC) is a single, threshold-independent metric for measuring the strength of the attack.  Figure~\ref{fig:roc} shows how we obtain the ROC curve.  AUC $ = 1$ implies that the attacker can correctly classify all target samples as members or non-members.
%

\noindent\textbf{Precision and Recall.} 
We set $\alpha=10\%$ as the false positive rate and choose the threshold accordingly, as shown in Fig.~\ref{fig:threshold}. 
For precision, we measure the percentage of samples correctly inferred as  members of the training set out of the total number of target samples inferred as members by the attack.
For recall, we measure the percentage of samples correctly inferred as members of the training set out of the total number of target samples that are actually members of the training set. 

\section{Results}\label{sec:results}

\begin{figure}[t]
    \centering
    
    \begin{subfigure}{0.38\textwidth}
     \includegraphics[width=\linewidth]{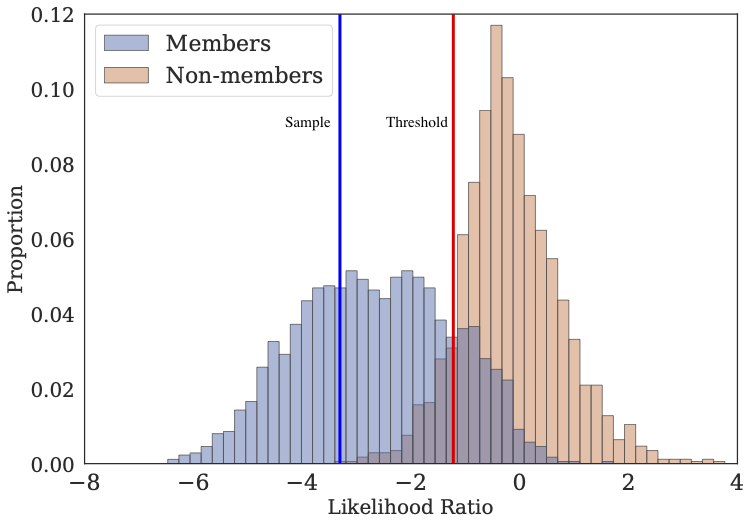}
     \footnotesize
     \caption{Likelihood Ratio $L(s)$ Histogram}
     \label{fig:hist:ours}
    \end{subfigure}
    \begin{subfigure}{0.38\textwidth}
     \includegraphics[width=\linewidth]{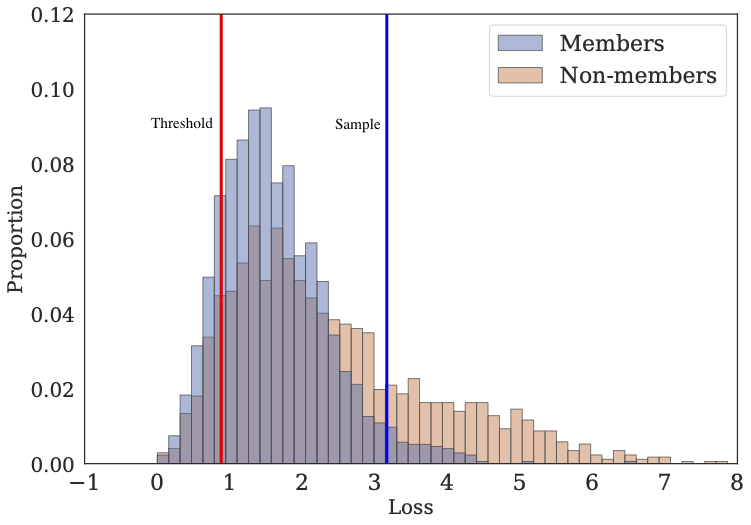}
    \footnotesize 
     \caption{Target Model Loss Histogram}
     \label{fig:hist:loss}
    \end{subfigure}
  \caption{(a) likelihood ratio histogram for training data members and non-members. (b) loss histogram for training data members and non-members. The blue lines in the two figures correspond to the same target sample (which is a random member of training-set). The red line is  is the threshold at $\alpha=10\%$ false positive rate. The threshold from our attack (a) is correctly able to label the test sample as a training-set member but the thresholds from the baseline attack (b) fails to do so.} 
    \label{fig:DPmodes}
    \vspace{-3ex}
\end{figure}

In this section, we discuss our experimental results and main observations. First, we explore the overall performance improvement of our approach over baselines. Later, we analyze the effectiveness of our approach across several factors of variation that have an effect on the leakage of the model. (e.g. length of samples, model size, including names etc.)
%
Finally, we explore correlations between samples that are deemed to be \emph{exposed} by our approach. We provide further studies and ablations on choosing a lower false-positive rate of $1\%$, using different energy formulation and changing target sequence lengths  in Appendix sections~\ref{app:low-fp},~\ref{app:energy}, and~\ref{app:seq-len}, respectively. 

\begin{table}[]
    \centering
    \caption{Overview of our attack on the ClinicalBERT-Base  model, using PubMed-BERT as the reference. Sample-level attack attempts to determine membership of a single sample, whereas patient-level determines membership of a patient based on all their notes. The MIMIC and i2b2 columns determine which dataset was used as non-members in the target sample pool.}
    \vspace{-2ex}
    \label{tab:overview}
    \begin{adjustbox}{width=0.99\linewidth, center}
     \newcolumntype{L}{>{\RaggedLeft\arraybackslash}p{0.06\linewidth}} 
  \newcolumntype{O}{>{\RaggedLeft\arraybackslash}m{0.07\linewidth}} 
  \newcolumntype{D}{>{\arraybackslash}m{0.15\linewidth}} 
  \newcolumntype{R}{>{\arraybackslash}m{0.29\linewidth}} 
\begin{tabular}{@{}clcccccSSSS@{}}

	\toprule
	& {\multirow{2}{*}{}} &  \multicolumn{2}{c}{{Sample-level}}&{}& \multicolumn{2}{c}{{Patient-level}}   \\
	\cmidrule{3-4} \cmidrule{6-7}
	&  Non-members &  {MIMIC}  &{i2b2}& {} & {MIMIC} & {i2b2}  \\
    \midrule
    \multirow{2}{*}{\STAB{AUC.}} 
    & (A) Model loss    & 0.662	&0.812&	&0.915&	1.000	\\
    & (B) Ours	        & 0.900	&0.881&	&0.992&	1.000  \\
           \midrule[0.1pt]
    \multirow{3}{*}{\STAB{Prec.}}
    & (A) w/ $\mu$ thresh.	& 61.5  &	77.6   & &	87.5    &	100.0   \\
    & (A) w/ Pop. thresh.   & 61.2  &	79.6   & &	87.5    &	92.5   \\
    &(B)  w/ Pop. thresh.	& 88.9  &	87.5    & &	93.4    &	92.5   \\
           \midrule[0.1pt]
    \multirow{3}{*}{\STAB{Rec.}}
    &(A) w/ $\mu$ thresh	&  55.7 &	55.8    &&	49.5	&   49.5  \\
    &(A) w/ Pop. thresh.    &  15.6 &	39.0    &&	49.5	&   100.0  \\
    &(B)  w/ Pop. thresh.	&  79.2	&    69.9    &&	100.0	&   100.0  \\
	\bottomrule
\end{tabular}

    \end{adjustbox}
    \vspace{-2ex}
\end{table}

\begin{figure}[]
    \centering
        \begin{subfigure}{0.37\textwidth}
     \includegraphics[width=\linewidth]{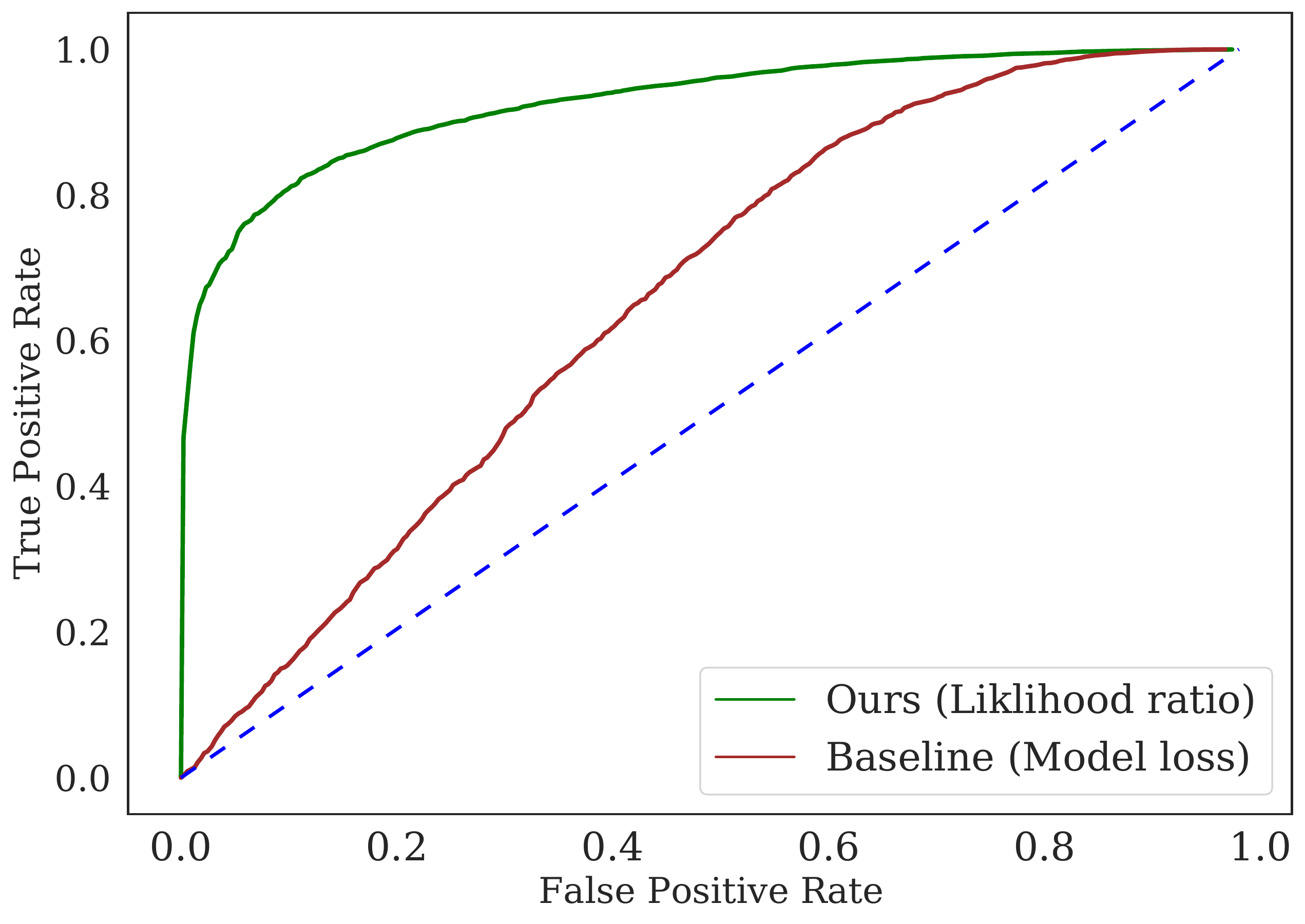}
    \footnotesize 
    \end{subfigure}
    \caption{The ROC curve of sample-level attack on Clinical-BERT with MIMIC used as non-member.  Green line shows our attack and  the red line shows the baseline loss-based attack. The blue dashed line shows AUC=$0.5$ (random guess). This figure corresponds to the results presented in the first column of Table~\ref{tab:overview}. }
    \label{fig:result:roc}
    \vspace{-2ex}
\end{figure}

\subsection{Comparison with Baseline}

Table~\ref{tab:overview}  shows the metrics for our attack and the baseline's on both sample and patient level, with held-out MIMIC-III and i2b2 medical notes used as non-member samples (Figure~\ref{fig:result:roc} shows the ROC curve).
%
The table shows that our method significantly outperforms the target model loss-based baselines~\cite{jagannatha2021membership,yeom2018privacy}, which threshold the loss of the target model based on either the mean of the training samples' loss ($\mu$), or the population samples' loss.  
Our attack's improvement over the baselines is more apparent in the case where both the members and non-members are from MIMIC-III. This case is harder for the baselines since members and non-members are much more similar and harder to distinguish if we only look at the loss of the target model. Our attack, however, is successful due to the use of a reference, which helps magnify the gap in the behavior of the target model towards members and non-members, thereby teasing apart similar samples.

 We can also see that in terms of precision/recall trade-off, our attack has a consistently higher recall, with an average higher precision. Population loss based thresholding ((A) w/ Pop. thresh.) has the lowest recall of $15.6\%$, which is due to members and non-members achieving similar losses from the target model due to their similarity. 
 This is also shown in Figure~\ref{fig:hist:loss}. 
 In Figure~\ref{fig:hist:ours}, however, we see a distinct separation between the member and non-member histogram distributions when we use the ratio statistic $L(s)$ as the test criterion for our attack. This results in the estimation of a useful threshold that correctly classifies the blue line sample as a member, as opposed to using only the target model loss (Figure~\ref{fig:hist:loss}). 
Finally, we observe that all the metrics have higher values on the patient-level attack, compared to sample-level, for both our attack and the baselines. This is due to the higher granularity of the patient level attack, as it makes the decision based on \emph{an aggregate} of multiple samples.

%
 %

\begin{table}[]
    \centering
    \caption{Effect of target sample length: Sample-level attack  on the ClinicalBERT-Base  model, using PubMed-BERT as the reference.  The MIMIC and i2b2 columns determine which dataset was used as non-members in the target sample pool. Short and long show a break down of the length of target samples. }
    \vspace{-2ex}
    \label{tab:sample-out-len}
    \begin{adjustbox}{width=0.99\linewidth, center}
     \newcolumntype{L}{>{\RaggedLeft\arraybackslash}p{0.06\linewidth}} 
  \newcolumntype{O}{>{\RaggedLeft\arraybackslash}m{0.07\linewidth}} 
  \newcolumntype{D}{>{\arraybackslash}m{0.15\linewidth}} 
  \newcolumntype{R}{>{\arraybackslash}m{0.29\linewidth}} 

\begin{tabular}{@{}cllllllclSSSS@{}}
	\toprule
	& {\multirow{2}{*}{}} &  \multicolumn{2}{c}{{Short}}&{}& \multicolumn{2}{c}{{Long}}   \\
	\cmidrule{3-4} \cmidrule{6-7}
	& Non-members  &  {MIMIC}  &{i2b2}& {} & {MIMIC} & {i2b2}  \\
    \midrule
    \multirow{2}{*}{\STAB{AUC.}} 
    &(A) Model loss     &0.516	&   0.756	&&   0.662	&   0.812  	\\
    &(B) Ours	        &0.830	&   0.845	&&   0.900	&   0.881   \\
    \midrule[0.1pt]
    \multirow{3}{*}{\STAB{Prec.}}
    &(A) w/ $\mu$ thresh.	& 50.9	&   70.0	&&   61.5	&   77.6   \\
    &(A) w/ Pop. thresh.    & 42.0	&   72.5	&&   61.2	&   79.6 \\ 
    &(B)  w/ Pop. thresh.   & 87.3	&   86.3	&&   88.9	&   87.5   \\
           \midrule[0.1pt]
    \multirow{3}{*}{\STAB{Rec.}}
    &(A) w/ $\mu$ thresh.	& 55.3	&   55.3	&&   55.7	&   55.8 \\
    &(A) w/ Pop. thresh.   &  7.2	&   26.3	&&   15.6	&   39.0\\
    &(B)  w/ Pop. thresh.	& 68.2	&   62.9	&&   79.2	&   69.9  \\
	\bottomrule
\end{tabular}

    \end{adjustbox}
        \vspace{-2ex}
\end{table}

\begin{table}[]
    \centering
    \caption{Effect of model size and training: Sample-level attack on the four different  ClinicalBERT models, using PubMed-BERT as the reference and the MIMIC data as non-members. Base++ (Large++) is same as Base (Large), but trained for more epochs. }
    \vspace{-2ex}
    \label{tab:sample-out-model}
    \begin{adjustbox}{width=0.99\linewidth, center}

\begin{tabular}{@{}clllll@{}}
	\toprule
	& Target Model  &  {Base} & {Base++} &  {Large} &{Large++}  \\
    \midrule
    \multirow{2}{*}{\STAB{AUC.}} 
&(A) Model loss     & 0.662	&   0.656	&   0.679	&   0.700	\\
&(B) Ours	        & 0.900	&   0.894	&   0.904	&   0.905 \\
\midrule[0.1pt]
\multirow{3}{*}{\STAB{Prec.}}
&(A) w/ $\mu$ thresh.   &61.5	&   61.0	&   62.4	&   64.9	   \\
&(A) w/ Pop. thresh.    &61.2	&   61.2	&   63.9	&   69.1	 \\ 
&(B)  w/ Pop. thresh.   &88.9	&   88.8	&   88.9	&   88.9	   \\
\midrule[0.1pt]
\multirow{3}{*}{\STAB{Rec.}}
&(A) w/ $\mu$ thresh.   &  55.7	&   55.8	&   56.4	&   56.1  \\
&(A) w/ Pop. thresh.    &  15.6	&   15.6	&   17.6	&   22.2 \\
&(B)  w/ Pop. thresh.   &  79.2	&   78.5	&   79.2	&   79.3   \\
	\bottomrule
\end{tabular}

    \end{adjustbox}
        \vspace{-2ex}
\end{table}

\vspace{-1ex}
\subsection{Effect of Sample Length and Model Size}\label{sec:model}
Tables~\ref{tab:sample-out-len} and~\ref{tab:sample-out-model} show the metrics for our attack and the baseline broken down based on the length of the target sample, and the size and training epochs of the target model, respectively. In Table~\ref{tab:sample-out-len}, the target model is same as that of Table~\ref{tab:overview}, ClinicalBERT-base. Short samples are those that have between $10$ to $20$ tokens, and long samples have $20$ to $60$ tokens.  We can see that both the baseline and our attacks show more leakage for long sentences than they do for short sequences, which could be due to the longer sentences being more unique and thus being more likely to provide a discriminative signal for a sequence-level decision.
%
Table~\ref{tab:sample-out-model} shows the attacks mounted on the four models from Table~\ref{tab:experiment_notation}. We see that leakage on all the models is very similar, however, the AUC on Large++ is consistently higher than on Base, which hints at the observation made by~\cite{carlini2021extracting} that larger models tend to have a higher capacity for memorization.



\subsection{Effect of Changing the Reference Model}
Table~\ref{tab:refs} studies how changing the reference model would affect the success of the attack. Here, \emph{Pubmed} is the reference model that is used in the previous experiments, and \emph{BERT-base} is Huggingface's pre-trained BERT. We observe that the attack using BERT-base performs well, but is  worse than using Pubmed, especially in terms of recall (true positive rate). The main reason behind this is the domain overlap between the Pubmed reference model and the model under attack. An ideal reference model for this attack would be trained on data from a domain that is similar to that of the target model's training set so as to better characterize the intrinsic complexity of the samples.
On the other hand, a reference model trained on a different data distribution (in this case Wikipedia) would give the same score to easy and difficult samples, thereby decreasing the true positive rate (recall), as shown in the table. 

\begin{table}[]
    \centering
    \caption{ Effect of reference model: Sample-level attacks on ClinicalBERT-Base model, using PubMed-BERT and standard bert-base-uncased as the reference and MIMIC data as non-member.  }
    \vspace{-2ex}
    \label{tab:refs}
    \begin{adjustbox}{width=0.99\linewidth, center}
     \newcolumntype{L}{>{\RaggedLeft\arraybackslash}p{0.06\linewidth}} 
  \newcolumntype{O}{>{\RaggedLeft\arraybackslash}m{0.07\linewidth}} 
  \newcolumntype{D}{>{\arraybackslash}m{0.15\linewidth}} 
  \newcolumntype{R}{>{\arraybackslash}m{0.29\linewidth}} 

\begin{tabular}{@{}cllllllclSSSS@{}}
	\toprule
	& {\multirow{2}{*}{}} &  \multicolumn{2}{c}{{Base}}&{}& \multicolumn{2}{c}{{Large++}}   \\
	\cmidrule{3-4} \cmidrule{6-7}
	&  Reference Model &  {pubmed}  &{bert}& {} & {pubmed} & {bert}  \\
    \midrule
    \multirow{2}{*}{\STAB{AUC.}} 
&(A) Model loss     & 0.662 &	0.662   &&	0.700	&   0.700  	\\
&(B) Ours	        & 0.900 &	0.883   &&	0.905	&   0.889  \\
\midrule[0.1pt]
\multirow{3}{*}{\STAB{Prec.}}
&(A) w/ $\mu$ thresh.&  61.5    &	61.5    &&  64.9	&   64.9 \\
&(A) w/ Pop. thresh. &  61.2    &	61.2    &&  69.1	&   69.1 \\ 
&(B)  w/ Pop. thresh.&  88.9    &	87.8    &&  88.9	&   88.0 \\
\midrule[0.1pt]
\multirow{3}{*}{\STAB{Rec.}}
&(A) w/ $\mu$ thresh.& 55.7&	55.7    &&	56.1	&   56.1  \\
&(A) w/ Pop. thresh. & 15.6&	15.6    &&	22.2	&   22.2 \\
&(B)  w/ Pop. thresh.& 79.2&	71.5    &&	79.3	&   72.6  \\
	\bottomrule
\end{tabular}
    \end{adjustbox}
    \vspace{-2ex}
\end{table}

\subsubsection{Effect of Inserting Names}\label{app:name-insertion}
Table~\ref{tab:overview-1b} shows results for attacking the name insertion model~\cite{lehman-etal-2021-bert}, shown as Base-b, where the patient's first and last name are prepended to each training sample.
%
%
We see that our attack's performance is better on the name-insertion model, compared to the base model, whereas the baseline attack performs worse (in the sample-level scenario). We hypothesize that this is due to the ``difficulty'' of the samples. Adding names to the beginning of each sample actually increases the entropy of the dataset overall, since in most cases they don't have a direct relation with the rest of the sentence (except for very few sentences that directly state a person's disease), therefore they might as well be random. 
This makes these sentences more difficult and harder to learn, as there is no easy pattern. Hence, on average, these sentences have higher loss values (2.14 for name inserted samples, vs. 1.61 for regular samples). However, for the non-members, since they don't have names attached to them, the average loss is the same (the $10\%$ FPR threshold is 1.32), and that is why the attack performs poorly on these samples, as most of the members get classified as non-members.
For our attack, since we use the reference, we are able to tease apart such hard samples as they are extremely less likely given the reference than they are given the target model.

\begin{table}[]
    \centering
    \caption{Effect of inserting names: Sample and Patient-level attacks on ClinicalBERT-Base and Base-b (name insertion)   model, using PubMed-BERT as the reference and MIMIC data as non-member. We study  the effect that inserting names into all training samples has on the leakage of the model.}
    \vspace{-2ex}
    \label{tab:overview-1b}
    \begin{adjustbox}{width=0.99\linewidth, center}
     \newcolumntype{L}{>{\RaggedLeft\arraybackslash}p{0.06\linewidth}} 
  \newcolumntype{O}{>{\RaggedLeft\arraybackslash}m{0.07\linewidth}} 
  \newcolumntype{D}{>{\arraybackslash}m{0.15\linewidth}} 
  \newcolumntype{R}{>{\arraybackslash}m{0.29\linewidth}} 

\begin{tabular}{@{}cllllllcSSSSS@{}}
	\toprule
	& {\multirow{2}{*}{}} &  \multicolumn{2}{c}{{Sample-level}}&{}& \multicolumn{2}{c}{{Patient-level}}   \\
	\cmidrule{3-4} \cmidrule{6-7}
	&  Target model &  {Base}  &{Base-b}& {} & {Base} & {Base-b}  \\
    \midrule
    \multirow{2}{*}{\STAB{AUC.}} 
&(A) Model loss     &  0.662	&   0.561	&&   0.915	&   0.953	\\
&(B) Ours	        &  0.900	&   0.960	&&   0.992	&   1.000 \\
\midrule[0.1pt]
\multirow{3}{*}{\STAB{Prec.}}
&(A) w/ $\mu$ thresh.   & 61.5  &	53.0    &&	87.5    &	100.0 \\
&(A) w/ Pop. thresh.    & 61.2  &	44.1    &&	87.5    &	91.1 \\ 
&(B)  w/ Pop. thresh.   & 88.9  &	90.2    &&	93.4    &	92.5 \\
\midrule[0.1pt]
\multirow{3}{*}{\STAB{Rec.}}
&(A) w/ $\mu$ thresh.	& 55.7&	54.0	&&49.5	    &48.5   \\
&(A) w/ Pop. thresh.    & 15.6&	7.8	    &&49.5	    &82.8      \\
&(B)  w/ Pop. thresh.   & 79.2&	91.3	&&100.0 &	100.0    \\
	\bottomrule
\end{tabular}

    \end{adjustbox}
\end{table}


\subsection{Correlations between Memorized Samples}

To evaluate whether there are correlations between samples that have high leakage based on our attack (i.e. training samples that are successfully detected as members), we conduct an experiment.
In this experiment, we create a new train and test dataset, by subsampling the main dataset and selecting $5505$ and $7461$ samples, respectively.  We label the training and test samples based on whether they are exposed or not, i.e. whether the attack successfully detects them as training samples or not, and get $2519$ and 3$283$ samples labeled as ``memorized'', for the train and test set. 
Since our goal is to see if we can find correlations between the memorized samples of the training set and use those to predict memorization on our test set, we create features for each sample, and then use those features with the labels to create a simple logistic regression classifier that predicts memorization. 

Table~\ref{tab:corr} shows these results in terms of precision and recall for predicting if a sample is ``memorized" or not, with different sets of features. The first $4$ rows correspond to individual handcrafted feature sets: (A) the number of digits in the sample, (B) length of a sample (in tokens), (C) the number of non-alphanumeric characters (this would be characters like '*', '-', etc.). (D) corresponds to feature sets that are obtained by encoding the tokenized sample by the frequency of each of its tokens, and then taking the 3 least frequent tokens' frequencies as features (the frequency comes from a frequency dictionary built on the training set). 
%
%
We can see that among the hand-crafted features, (C) is most indicative, as it counts the characters that are more out-of-distribution and are possibly not determined by grammatical rules or consistent patterns. 
(C) and (D) concatenated together perform slightly better than (C) alone,  which could hint at the effect frequency of tokens and how common they are could have on memorization. We also get a small improvement over these by concatenating (B), (C), and (D),  which shows the length has a slight correlation too. 

%

\begin{table}[]
    \centering
    \caption{Analysis of correlations between samples that are leaked through our attack. We want to see what features are shared among all leaked samples by extracting a list of possible features and training a simple logistic regression model on a subset of the original training data ($D$), and then testing it on another subset.  The logistic regression model tries to predict whether a sample would be leaked or not (based on whether our model has classified it as a member or not). The precision and recall here are those of the logistic regression model, for predicting leaked training samples.  }
    \vspace{-2ex}
    \label{tab:corr}
    \begin{adjustbox}{width=1.0\linewidth, center}
     \newcolumntype{L}{>{\RaggedLeft\arraybackslash}p{0.06\textwidth}} 
  \newcolumntype{O}{>{\RaggedLeft\arraybackslash}m{0.07\textwidth}} 
  \newcolumntype{D}{>{\arraybackslash}m{0.15\textwidth}} 
  \newcolumntype{R}{>{\arraybackslash}m{0.29\textwidth}} 
\begin{tabular}{@{}llSSSSScSSSSS@{}}
	\toprule
	 {\multirow{2}{*}{Features}} &  \multicolumn{2}{c}{{Train}}&{ }& \multicolumn{2}{c}{{Test}}  \\
	\cmidrule{2-3} \cmidrule{5-6} 
	    & {Prec.} & {Rec.} &  &{Prec.} & {Rec.}    \\
    \midrule
    
 (A) \#Digits      &0.0    &	    0.0     &   &	0.0	    &   0.0     \\
 (B) Seq. Len      &0.0    &	    0.0     &   &	0.0	    &   0.0     \\
 (C) \#Non-alphanumeric      &71.2   &   	46.6    &   &	69.2    &	47.5    \\
 (D) 3 Least Frequent        &68.9   &   	40.5    &   &	63.8    &	39.2    \\
 (C) \& (D)     &73.9   &   	58.8    &   &	71.1    &	57.8    \\
 (B) \& (C) \& (D)      &74.3   &   	61.3    &   &	72.1    &	61.3    \\
 (A) \& (B) \& (C) \& (D)      &74.3   &   	61.3    &   &	72.1    &	61.3    \\
  
	\bottomrule
\end{tabular}

    \end{adjustbox}
    \vspace{-2ex}
\end{table}

\section{Related Work}
 \label{sec:privacy_violations_lm}

%
Prior work on measuring memorization and leakage in machine learning models can be classified into two main categories: (1)~membership inference attacks and (2)~training data extraction attacks.%
%
\paragraph{Membership inference.}

Membership Inference Attacks (MIA) try to determine whether or not a target sample was used in training a target model~\cite{shokri2017membership,yeom2018privacy}. 
These attacks can be seen as privacy risk analysis tools~\cite{murakonda2020ml, nasr2021adversary, kandpal2022deduplicating}, which help reveal how much the model has memorized the individual samples in its training set, and what the risk of individual users is~\cite{ nasr2019comprehensive, memberinf2, memberinf3, ye2021enhanced, carlini2021membership}.
A group of these attacks rely on behavior of shadow models  to determine the membership of given samples~\cite{jayaraman2021revisiting,shokri2017membership}. 
%
%
%
~\citeauthor{songaudit} mounts such an attack on LSTM-based text-generation models, ~\citeauthor{mahloujifar2021membership} mounts one on word embedding,~\citeauthor{10.1162/tacl_a_00299} applies it to machine translation and more recently,~\citeauthor{shejwalkar2021membership} mounts it on transformer-based NLP classification models. 
Mounting such attacks is usually costly, as their success relies upon training multiple shadow models on different partitionings of shadow data, and access to adequate shadow data for training such models. 

Another group of MIAs relies solely on the loss value of the target sample, under the target model, and thresholds this loss to determine membership~\cite{jagannatha2021membership,yeom2018privacy}.
~\citeauthor{song2020information} mount such an attack on word embedding, where they try to infer if given samples were used in training different embedding models.~\citet{jagannatha2021membership}, which is the work closest to ours, uses a thresholding loss-based attack to infer membership on MLMs. 
Our approach instead incorporates a reference model by using an energy-based formulation to mount a likelihood ratio based attack and achieves higher AUC as shown in the results.


\paragraph{Training data extraction.}
Training data extraction quantifies the risk of extracting training data by probing a trained language model~\cite{saleme2020, carlini2019secret, santiago-snapshot-2020, carlini2021extracting,carlini2022quantifying,nakamura2020kart}.  
One such prominent attacks on NLP models is that of~\citet{carlini2021extracting}, where they take more than half a million samples from different GPT-2 models, sift through the samples using a membership inference method to find samples that are most likely to have been memorized.
~\citet{lehman-etal-2021-bert} mount the same data extraction attack on MLMs, but their results are inconclusive as to how much MLMs memorize samples.
They also mount other types of attacks, where they try to extract a person's name given their disease, or disease given name, but in all their attacks, they only use signals from the target model and consistently find that a frequency-based baseline (i.e. one that would always guess the most frequent name/disease) is more successful. 

\section{Conclusions}
In this paper, we introduce a principled membership inference attack based on likelihood ratio testing to measure the training data leakage of Masked Language  Models (MLMs). In contrast to prior work on MLMs, we rely on signals from both the model under attack and a reference model to decide the membership of a sample. This enables performing successful membership inference attacks on data points that are hard to fit, and therefore cannot be detected using the prior work.  We also perform an analysis of \textit{why} these models leak, and \text{which} data points are more susceptible to memorization. Our attack shows that MLMs are significantly prone to memorization. This work calls for designing robust privacy mitigation algorithms for such language models. 

\section*{Limitations}

Membership inference attacks form the foundation of privacy auditing and memorization analysis in machine learning. As we show in this paper, and as it is shown in the recent work~\cite{carlini2021membership,ye2021enhanced}, these attacks are very efficient in identifying privacy vulnerabilities of models with respect to individual data records. However, for a thorough analysis of data privacy, it is not enough to rely only on membership inference attacks. We thus would need to extend our analysis to reconstruction attacks and property inference attacks.


\section*{Ethics Statement}

We use two datasets in this paper, MIMIC-III and i2b2, both of which contain sensitive data and can only be accessed by request\footnote{Access can be requested through~\url{https://mimic.mit.edu/docs/gettingstarted/} and~\url{https://portal.dbmi.hms.harvard.edu/projects/n2c2-nlp/}} and after agreeing to the data usage and confidentiality terms\footnote{Data Usage Agreements (DUA) are available in~\url{https://physionet.org/content/mimiciii/view-dua/1.4/} and ~\url{https://projects.iq.harvard.edu/files/n2c2/files/n2c2_data_sets_dua_preview_-_academic_user.pdf} for the datasets, respectively.} and passing proper training for ethical and privacy-preserving use of the data.
For reproduction of our results, code will be made available only by request and for research purposes, only to researchers who provide proof of authorized access to the datasets (by forwarding the access granted emails from MIMIC-III and i2b2 to the first author\footnote{fmireshg@eng.ucsd.edu}).

To protect models against membership inference attacks, like the one proposed in this work, differentially private training algorithms~\cite{abadi2016deep,chaudhuri2011differentially} can be used, as they are theoretically designed to protect the membership of each data record individually. Other methods such as adversarial training~\cite{mireshghallah2021privacy} and personally identifiable information scrubbing~\cite{dernoncourt2017identification} can also be used, however, they do not provide the worst-case guarantees that differential privacy does~\cite{brown2022does}.

\section*{Acknowledgements}
The authors would like to thank the anonymous reviewers and meta-reviewers for their helpful feedback. We also thank our colleagues at the UCSD Berg Lab and NUS for their helpful comments and feedback. 

%
%


\bibliography{anthology,acl_latex}
\bibliographystyle{acl_natbib}

\appendix
\clearpage
\label{sec:appendix}

\section{Appendix}

\begin{table}[]
    \centering
    \caption{Summary of our attack's notations.}
    \vspace{-2ex}
    \label{tab:attack_notation}
    \begin{adjustbox}{width=0.95
    \linewidth, center}

\begin{tabular}{cllll}
	\toprule
	&Notation &  {Explanation} \\
    \midrule
    \multirow{6}{*}{\STAB{}} 
&   D	    &   Training dataset    \\
&   S	    &   Set of target samples whose membership we want to determine.  \\
&   s	    &   A given target sample ($s\in S$, $p(s\in D)=0.5$)) \\
&   R	    &   Reference model \\
&   t	    &   Attack threshold    \\
&   L(s)	&   Likelihood ratio for given sample $s$   \\
	\bottomrule
\end{tabular}

    \end{adjustbox}
\end{table}

\subsection{Notations}\label{app:notation}
To summarize and clarify the notations used in the paper for explaining our attack, we added Table~\ref{tab:attack_notation}.

\subsection{Detailed Experimental Setup}\label{app:setup}

\subsubsection{Code and Data Access}\label{sec:app:data}
We use two datasets in this paper, MIMIC-III and i2b2, both of which contain sensitive data and can only be accessed by request through~\url{https://mimic.mit.edu/docs/gettingstarted/} and~\url{https://portal.dbmi.hms.harvard.edu/projects/n2c2-nlp/}, and after agreeing to the data usage and confidentiality terms\footnote{Data Usage Agreements (DUA) are available in~\url{https://physionet.org/content/mimiciii/view-dua/1.4/} and ~\url{https://projects.iq.harvard.edu/files/n2c2/files/n2c2_data_sets_dua_preview_-_academic_user.pdf} for the datasets, respectively.} and passing proper training for ethical and privacy-preserving use of the data.
For reproduction of our results, code will be made available only by request and for research purposes, only to researchers who provide proof of authorized access to the datasets (by forwarding the access granted emails from MIMIC-III and i2b2 to the first author\footnote{fmireshg@eng.ucsd.edu}).

\subsubsection{Datasets}\label{app:datasets}

We run our attack on two sets of target samples, one we denote by ``MIMIC'' and the other by ``i2b2'' in the results (both medical notes). For both of these, the ``members'' portion of the target samples is from the training set ($D$) of our target models, which is the MIMIC-III dataset. However, the non-members are different. For the results shown under MIMIC, the non-members are a held-out subset of the MIMIC data that was not used in training. For i2b2, the non-members are from a different (but similar) dataset, i2b2. Below we elaborate on this setup and each of these datasets. 

\paragraph{MIMIC-III} The target models we attack are trained (by \citeauthor{lehman-etal-2021-bert}) on the pseudo re-identified MIMIC III notes which consist of $1,247,291$  electronic health records (EHR) of  $46,520$ patients. These records have been altered such that the original first and last names are replaced with new first and last names sampled from the US Census data.  Only $27,906$ of these patients had their names explicitly mentioned in the EHR.

For the attack's target data, we use a held-out subset of MIMIC-III consisting of $89$ patients ($4072$ sample sequences) whose data was not used during training of the ClinicalBERT target models and use them as ``non-member''  samples. For ``member'' samples, we take a $99$ patient subsample of the entire training data, and then subsample $4072$ sample sequences from that (we do this $10$ times for each attack and average the results over), so that the number of member and non-member samples are the same and the target pool is balanced.

\paragraph{i2b2} This dataset was curated for the i2b2 de-identification of protected health information (PHI) challenge in 2014~\cite{i2b2}. 
We use this dataset as a secondary non-member dataset, since it is similar in domain to MIMIC-III (both are medical notes), is larger in terms of size than the held-out MIMIC-III set,  and has not been used as training data for our models.
We subsample $99$ patients from i2b2, consisting of $18561$ sequences, and use them as non-members.

The population data that we use to evaluate the distribution of likelihood ratio over the null hypothesis (which is used to compute the threshold) is disjoint with the non-member set that we use to evaluate the attack.  We randomly select $99$ patients from the i2b2 dataset for this purpose.

\subsubsection{Target Models}\label{app:target}

We perform our attack on $5$ different pre-trained models, that are all trained on MIMIC-III, but with different training procedures:

\paragraph{ClinicalBERT (Base)} BERT-base architecture trained over the pseudo re-identified MIMIC-III notes  for 300k iterations for sequence length 128 and for 100k iterations for sequence length 512.

\paragraph{ClinicalBERT++ (Base++)} BERT-base architecture  trained over the pseudo re-identified MIMIC III notes for 1M iterations at a sequence length of 128.

\paragraph{ClinicalBERT-Large (Large).} BERT-large architecture trained over the pseudo re-identified MIMIC-III notes  for 300k iterations for sequence length 128 and for 100k iterations for sequence length 512. 

\paragraph{ClinicalBERT-large++ (Large++)} BERT-large architecture  trained over the pseudo re-identified MIMIC III notes for 1M iterations at a sequence length of 128.

\paragraph{ClinicalBERT-b ((Base-b), Name Insertion).} Same training and architecture as ClincalBERT-base, but that the patient’s surrogate name is prepended to the beginning of every sentence. This model is used to 
identify the effect of name-insertion on the memorization of BERT-based models~\cite{lehman-etal-2021-bert}.

\subsubsection{Computational Resources}
For this paper, we did not train any models, so the GPU training time is $0$ hours. However, for getting the likelihoods, we ran inference on the sequences in our target samples pool. For that, we used an RTX2080 GPU with 11GB of memory for  18 hours.

\subsection{Further Studies}\label{app:further}

\begin{table*}[]
    \centering
    \caption{Sample-level attack results (with \textbf{$\alpha=10\%$} false positive rate used for thresholding) on the four  ClinicalBERT models, plus the Base-b model (name insertion). We use PubMed-BERT as the reference model and the MIMIC data as non-members. Base++ (Large++) is same as Base (Large), but trained for more epochs. Base-b is the same as the Base model, but the training data was modified to prepend patient's first and last name to each sequence. This table studies effect of model size, training and sequence length on leakage.}
    \vspace{-2ex}
    \label{tab:sample-out-fp10-all}
    \begin{adjustbox}{width=0.99\linewidth, center}
     \newcolumntype{L}{>{\RaggedLeft\arraybackslash}p{0.06\linewidth}} 
  \newcolumntype{O}{>{\RaggedLeft\arraybackslash}m{0.07\linewidth}} 
  \newcolumntype{D}{>{\arraybackslash}m{0.15\linewidth}} 
  \newcolumntype{R}{>{\arraybackslash}m{0.29\linewidth}} 

\begin{tabular}{@{}cllllllclSSSSSSSSSS@{}}
	\toprule
	& {\multirow{2}{*}{}} &  \multicolumn{5}{c}{{Short}}&{}& \multicolumn{5}{c}{{Long}}   \\
	\cmidrule{3-7} \cmidrule{9-14}
	& Non-members  &  {Base}  &{Base++}& {Large}  &{Large++}&{Base-b}& {} &  {Base}  &{Base++}& {Large}  &{Large++}&{Base-b}&{} \\
    \midrule
    \multirow{2}{*}{\STAB{AUC.}} 
    &(A) Model loss     &0.662	&0.656	&0.679	&0.700&	0.561&& 0.516	&0.513	&0.509&	0.536&	0.391 	\\
    &(B) Ours	        &0.900	&0.894	&0.904	&0.905&	0.960&& 0.830	&0.827	&0.835&	0.843&	0.912  \\
    \midrule[0.1pt]
    \multirow{3}{*}{\STAB{Prec.}}
    &(A) w/ $\mu$ thresh.	&61.5	&61.0	&62.4	&64.9	&53.0	&&50.9	&50.6	&50.7	&52.5	&44.3   \\
    &(A) w/ Pop. thresh.    &61.2	&61.2	&63.9	&69.1	&44.1	&&42.0	&40.6	&40.0	&43.4	&25.9 \\ 
    &(B)  w/ Pop. thresh.   &88.9	&88.8	&88.9	&88.9	&90.2	&&87.3	&87.3	&87.3	&87.3	&89.3   \\
           \midrule[0.1pt]
    \multirow{3}{*}{\STAB{Rec.}}
    &(A) w/ $\mu$ thresh.   &55.7&	55.8&	56.4&	56.1&	54.0&&	55.3	&55.1	&56.2	&55.8	&54.3  \\
    &(A) w/ Pop. thresh.    &15.6&	15.6&	17.6&	22.2&	7.8	&&  7.2	    &6.8	&6.6	&7.6	&3.5    \\
    &(B)  w/ Pop. thresh.   &79.2&	78.5&	79.2&	79.3&	91.3&&	68.2	&68.2   &68.4	&68.4	&82.7   \\
	\bottomrule
\end{tabular}

    \end{adjustbox}
\end{table*}

\subsubsection{Lower False Positive Rate }\label{app:low-fp}
All the threshold-dependant results (precision and recalls) in Section~\ref{sec:results}  are reported with a threshold set for having~\textbf{$\alpha=10\%$} false positive rate (using the mechanism shown in Figure~\ref{fig:threshold}). In this section, we want to look at lower false positive rates, like we do in Figure~\ref{fig:result:roc-log}, and see how well our attack does when precision is very important to us and we do not want to get any false positives.  These results are shown in Table~\ref{tab:sample-out-fp1-all}. (This table corresponds with Table~\ref{tab:sample-out-model} from Section~\ref{sec:model} and the sample-level part of Table~\ref{tab:overview-1b}.) We can see that compared to Table~\ref{tab:sample-out-fp1-all}, as we are decreasing the false positive rate, the performance gap between our attack and the baseline increases drastically, showin that our attack performs really well under tight false positive rate constraints. 
Note that AUC is not threshold dependant therefore it has the same value in both tables. 

\begin{table*}[]
    \centering
    \caption{Sample-level attack results (with \textbf{$\alpha=1\%$} false positive rate used for thresholding) on the four  ClinicalBERT models, plus the Base-b model (name insertion). We use PubMed-BERT as the reference model and the MIMIC data as non-members. Base++ (Large++) is same as Base (Large), but trained for more epochs. Base-b is the same as the Base model, but the training data was modified to prepend patient's first and last name to each sequence. This table studies effect of model size, training and sequence length on leakage.}
    \vspace{-2ex}
    \label{tab:sample-out-fp1-all}
    \begin{adjustbox}{width=0.99\linewidth, center}
     \newcolumntype{L}{>{\RaggedLeft\arraybackslash}p{0.06\linewidth}} 
  \newcolumntype{O}{>{\RaggedLeft\arraybackslash}m{0.07\linewidth}} 
  \newcolumntype{D}{>{\arraybackslash}m{0.15\linewidth}} 
  \newcolumntype{R}{>{\arraybackslash}m{0.29\linewidth}} 

\begin{tabular}{@{}cllllllclSSSSSSSSSS@{}}
	\toprule
	& {\multirow{2}{*}{}} &  \multicolumn{5}{c}{{Short}}&{}& \multicolumn{5}{c}{{Long}}   \\
	\cmidrule{3-7} \cmidrule{9-14}
	& Non-members  &  {Base}  &{Base++}& {Large}  &{Large++}&{Base-b}& {} &  {Base}  &{Base++}& {Large}  &{Large++}&{Base-b}&{} \\
    \midrule
    \multirow{2}{*}{\STAB{AUC.}} 
    &(A) Model loss     &0.662	&0.656	&0.679	&0.700	&0.561	&&0.516	&0.513	&0.509	&0.536	&0.391 	\\
    &(B) Ours	        &0.900	&0.894	&0.904	&0.905	&0.960	&&0.830	&0.827	&0.835	&0.843	&0.912  \\
    \midrule[0.1pt]
    \multirow{3}{*}{\STAB{Prec.}}
    &(A) w/ $\mu$ thresh.	&61.5	&61.0	&62.4	&64.9	&53.0	&&50.9	&50.6	&50.7	&52.5	&44.3   \\
    &(A) w/ Pop. thresh.    &53.7	&53.8	&62.1	&71.4	&35.1	&&9.4	&5.8	&9.3	&18.3	&7.5 \\ 
    &(B)  w/ Pop. thresh.   &98.5	&98.4	&98.4	&98.4	&98.8	&&97.9	&97.9	&98.0	&97.9	&98.4   \\
           \midrule[0.1pt]
    \multirow{3}{*}{\STAB{Rec.}}
    &(A) w/ $\mu$ thresh.   &55.7	&55.8	&56.4	&56.1	&54.0	&&55.3	&55.1	&56.2	&55.8	&54.3  \\
    &(A) w/ Pop. thresh.    &1.1	&1.1	&1.6	&2.4	&0.5	&&0.1	&0.1	&0.1	&0.2	&0.1   \\
    &(B)  w/ Pop. thresh.   &60.4	&58.6	&57.6	&56.2	&78.3	&&43.0	&43.1	&45.2	&42.9	&58.7   \\
	\bottomrule
\end{tabular}

    \end{adjustbox}
\end{table*}

\subsubsection{Using  Normalized Energy }\label{app:energy}
In the paper, we use $E(s;\theta)$, as shown in Equation~\ref{eq:energy} for finding the likelihood ratio. In other words, for finding the likelihood ratio, we basically calculate the loss of the target model and reference model (using $15\%$ masking and averaging over 10 times)  on the given sequence $s$, and subtract them. 
However, another way to approach this problem of calculating likelihood ratio is to use the normalized energy (instead of the loss) as introduced in~\cite{goyal2022exposing}, instead of the loss. For calculating the normalized energy, we mask each token in the sequence, one token at a time (instead of $15\%$), calculated the loss, and average over all the tokens. To see how this energy does, we have used it to mount our attack, and we show the results in Table~\ref{tab:sample-out-fp10-all-energy}. 

Compared to using loss (Table~\ref{tab:sample-out-fp10-all}) it seems like the AUC for normalized energy is higher overall, 
\begin{table*}[]
    \centering
    \caption{Sample-level attack results (with \textbf{$\alpha=10\%$} false positive rate used for thresholding) on the four  ClinicalBERT models, plus the Base-b model (name insertion). Here we use ``normalized energy'' as a proxy for likelihood, instead of using the 15\% masked energy formulation of Equation~\ref{eq:energy}. We use PubMed-BERT as the reference model and the MIMIC data as non-members. Base++ (Large++) is same as Base (Large), but trained for more epochs. Base-b is the same as the Base model, but the training data was modified to prepend patient's first and last name to each sequence. This table studies effect of model size, training and sequence length on leakage.}
    \vspace{-2ex}
    \label{tab:sample-out-fp10-all-energy}
    \begin{adjustbox}{width=0.99\linewidth, center}
     \newcolumntype{L}{>{\RaggedLeft\arraybackslash}p{0.06\linewidth}} 
  \newcolumntype{O}{>{\RaggedLeft\arraybackslash}m{0.07\linewidth}} 
  \newcolumntype{D}{>{\arraybackslash}m{0.15\linewidth}} 
  \newcolumntype{R}{>{\arraybackslash}m{0.29\linewidth}} 

\begin{tabular}{@{}cllllllclSSSSSSSSSS@{}}
	\toprule
	& {\multirow{2}{*}{}} &  \multicolumn{5}{c}{{Short}}&{}& \multicolumn{5}{c}{{Long}}   \\
	\cmidrule{3-7} \cmidrule{9-14}
	& Non-members  &  {Base}  &{Base++}& {Large}  &{Large++}&{Base-b}& {} &  {Base}  &{Base++}& {Large}  &{Large++}&{Base-b}&{} \\
    \midrule
    \multirow{2}{*}{\STAB{AUC.}} 
    &(A) Model loss     &0.685	&0.675	&0.686	&0.714	&0.547	&&0.487	&0.482	&0.476	&0.522	&0.329	\\
    &(B) Ours	        &0.916	&0.910	&0.917	&0.924	&0.972	&&0.871	&0.869	&0.873	&0.881	&0.950 \\
    \midrule[0.1pt]
    \multirow{3}{*}{\STAB{Prec.}}
    &(A) w/ $\mu$ thresh.	&63.02&	62.18&	62.43&	63.70&	53.00&&	49.24&	49.23&	48.78&	50.44&	42.58   \\
    &(A) w/ Pop. thresh.    &64.38&	63.94&	66.78&	72.47&	34.68&&	36.55&	35.01&	36.49&	50.65&	7.74 \\ 
    &(B)  w/ Pop. thresh.   &89.07&	89.05&	89.09&	89.21&	90.40&&	88.07&	88.10&	88.15&	88.19&	89.97   \\
           \midrule[0.1pt]
    \multirow{3}{*}{\STAB{Rec.}}
    &(A) w/ $\mu$ thresh.   &58.11	&58.27	&58.57	&58.77	&55.29	&&59.17	&59.13	&59.70	&59.72	&55.42  \\
    &(A) w/ Pop. thresh.    &17.90	&17.57	&19.91	&25.90	&5.26	&&5.75	&5.37	&5.73	&10.22	&0.84   \\
    &(B)  w/ Pop. thresh.   &80.66	&80.48	&80.82	&81.82	&93.23	&&73.40	&73.61	&73.98	&74.21	&89.21   \\
	\bottomrule
\end{tabular}

    \end{adjustbox}
\end{table*}

\subsubsection{Results for Short Sequences}\label{app:seq-len}
All the results in Section~\ref{sec:results} are reported for long sequences (more than 20 tokens long), except those reported in Table~\ref{tab:sample-out-len}, where we ablate the privacy risks for sequences of different lengths. In this section, for the sake of completion, we are reporting results for short sequences as well as long sequences, for all the five models we study. These results are shown in Table~\ref{tab:sample-out-fp10-all}. (This table corresponds with Table~\ref{tab:sample-out-model} from Section~\ref{sec:model} and the sample-level part of Table~\ref{tab:overview-1b}.)

\subsubsection{Qualitative Comparison with Baseline}

Figures~\ref{fig:hist:ours} and~\ref{fig:hist:loss}  show histogram visualizations of $L(s)$ (likelihood ratio statistic) and model loss, over  the target sample pool (i.e. mixture of members and non-members from the MIMIC dataset), respectively. Here the target model is ClinicalBERT-Base.
The blue line represents a target query sample drawn from the member set.
The point of these visualizations is to show a case in which a sample is misclassified as a non-member by the model loss baseline, but is correctly classified as a member using our attack. 

The member and non-member distributions' histograms are shown via light blue and light orange bars respective. The red line indicates the threshold that is selected such that $\alpha=0.1$, i.e. $10\%$ false positive rate.  
In Figure~\ref{fig:hist:ours} we see a distinct separation between the member and non-member histogram distributions when we use $L(s)$ as the test criterion for our attack. This results in the estimation of a useful threshold that correctly classifies the blue line sample as a member. 
In contrast, the baseline attack by~\cite{jagannatha2021membership}, in Figure~\ref{fig:hist:loss} based solely upon the target model's loss leads to a high overlap between the member and non-member histograms which leads to a threshold that misclassifies the sample represented by the blue line.
These histograms show that the reference model used in our method helps in getting a sense of how hard each sample is in general, and puts each point in perspective.

\subsubsection{ROC Curve Magnified}\label{app:ROC}

\begin{figure}[h!]
    \centering
        \begin{subfigure}{0.49\textwidth}
     \includegraphics[width=\linewidth]{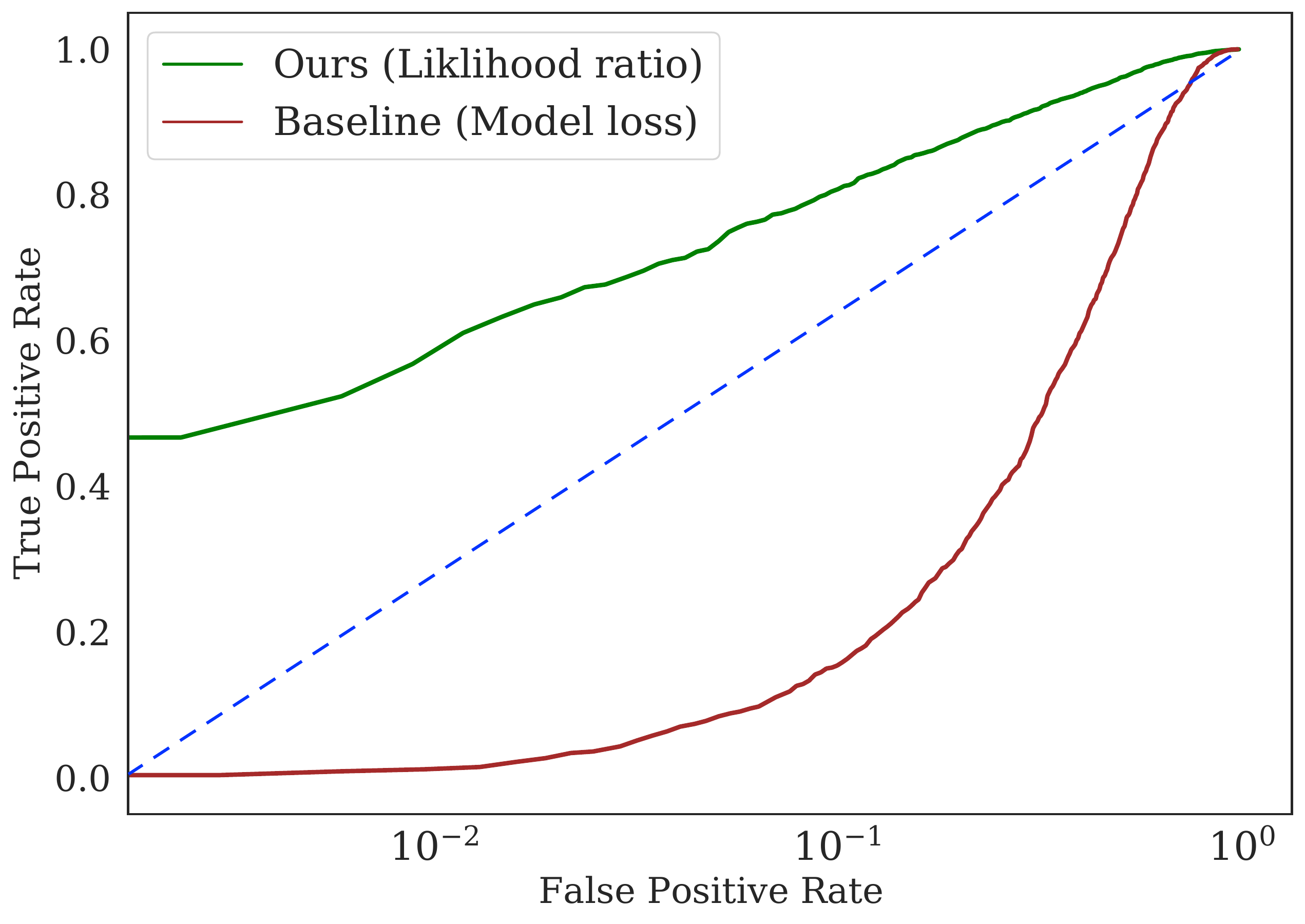}
    \footnotesize 
    \end{subfigure}
    \caption{The ROC curve of sample-level attack, with logarithmic scale x-axis. performed on Clinical-BERT using MIMIC samples as non-members.  Green line shows our attack and  the red line shows the baseline loss-based attack. The blue dashed line shows AUC=$0.5$ (random uess). We find false positive and true positive rates over the target sample pool, which is comprised of members and non-members. This curve corresponds to Table~\ref{tab:overview} in the results, and is the zoomed-in version of Figure~\ref{fig:result:roc}.}
    \label{fig:result:roc-log}
    \vspace{-2ex}
\end{figure}

In Figure~\ref{fig:result:roc-log} We have  plotted Figure~\ref{fig:result:roc} from the results, but with logarithmic x-axis, to zoom in on the low false positive rate section and really show the differences between our attack and the baseline.  

\subsection{Extended Related Works}
 \label{sec:privacy_violations_lm}

Since our work proposes an attack for quantifying leakage of masked language models (MLMs), based on the likelihood ratio, there are two lines of work that are related to ours: (1)~work surrounding attacks/leakage on machine learning models (2)~work on calculating sequence likelihood for MLMs.  Prior work on measuring memorization and leakage in machine learning models, and specifically NLP models can itself be classified into two main categories: (1)~membership inference attacks and (2)~training data extraction attacks. Below we discuss each line of work in more detail. 
%
\paragraph{Membership inference.}

Membership Inference Attacks (MIA) try to determine whether or not a target sample was used in training a target model~\cite{shokri2017membership,yeom2018privacy}. 
These attacks be seen as privacy risk analysis tools~\cite{murakonda2020ml, nasr2021adversary, kandpal2022deduplicating}, which help reveal how much the model has memorized the individual samples in its training set, and what the risk of individual users is~\cite{ nasr2019comprehensive, memberinf2, memberinf3, ye2021enhanced, carlini2021membership}
A group of these attacks rely on behavior of shadow models (models trained on data similar to training, to mimic the target model) to determine the membership of given samples~\cite{jayaraman2021revisiting,shokri2017membership}. 
In the shadow model training procedure the adversary trains a batch of models $m_1, m_2, . . . , m_k$ as shadow models, with data from the target user. Then, it trains  $m_1',m_2' . . . , m_k'$ without the data from the target user and then tries to find some statistical disparity between these models~\cite{mahloujifar2021membership}.
Shadow-based attacks have been mounted on NLP models as well:~\cite{songaudit} mounts such an attack on LSTM-based text-generation models, ~\cite{mahloujifar2021membership} mounts one on word embedding,~\cite{10.1162/tacl_a_00299} applies it to machine translation and more recently,~\cite{shejwalkar2021membership} mounts it on transformer-based NLP classification models. 
Mounting such attacks is usually costly, as their success relies upon training multiple shadow models, and access to adequate shadow data for training such models. 

Another group of MIAs relies solely on the loss value of the target sample, under the target model, and thresholds this loss to determine membership~\cite{jagannatha2021membership,yeom2018privacy}.
~\citeauthor{song2020information} mount such an attack on word embedding, where they try to infer if given samples were used in training different embedding models.~\citeauthor{jagannatha2021membership}, which is the work closest to ours, uses a thresholding loss-based attack to infer membership on MLMs. 
Although our proposed attack is also a threshold-based one, it is different from prior work by: (a)~applying likelihood ratio testing using a reference model and (b)~calculating the likelihood through our energy function formulation. These two components cause our attack to have higher AUC, as shown in the results.

We refer the reader to the framework introduced by~\cite{ye2021enhanced} that formalizes different membership inference attacks and compares their performance on benchmark ML tasks.

\paragraph{Training data extraction.}
Training data extraction quantifies the risk of extracting training data by probing a trained language model~\cite{saleme2020, carlini2019secret, santiago-snapshot-2020, carlini2021extracting,carlini2022quantifying,nakamura2020kart}.  
The most prominent of such attacks, on NLP models is that of~\citet{carlini2021extracting}, where they take more than half a million samples from different GPT-2 models,  sift through the samples using a membership inference method to find samples that are more likely to have been memorized, and finally, once they have narrowed down the samples to 1800, they check the web to see if such samples might have been in the GPT-2 training set. They find that over 600 of those 1800 samples were verbatim training samples.~\citet{lehman-etal-2021-bert} mount the same data extraction attack on MLMs, but their results are somehow inconclusive as to how much MLMs memorize samples, as only 4\% of generated sentences with a patient's name also contain one of their true medical conditions. They also mount other type of attacks, where they try to extract a person's name given their disease, or disease given name, but in all their attacks, they only use signals from the target model and consistently find that a frequency-based baseline (i.e. one that would always guess the most frequent name/disease) is more successful. 

\end{document}